\documentclass[runningheads]{llncs}
\usepackage[T1]{fontenc}
\usepackage{graphicx}
\usepackage{amsmath}
\usepackage{hyperref}
\usepackage{booktabs}
\usepackage{amssymb}
\usepackage{array}

\usepackage{color}

\begin{document}
\title{ICPR 2024 Competition on Safe Segmentation of Drive Scenes in Unstructured Traffic and Adverse Weather Conditions} 
\titlerunning{IDD-AW Safe Segmentation Competition}
%
\author{Furqan Ahmed Shaik\inst{1}\thanks{furqan.shaik@research.iiit.ac.in} \and
Sandeep Nagar\inst{1}  \and Aiswarya Maturi\inst{2} \and Harshit Kumar Sankhla \inst{3} \and Dibyendu Ghosh \inst{3} \and Anshuman Majumdar \inst{3} \and Srikanth Vidapanakal \inst{3}  \and Kunal Chaudhary\inst{4} \and Sunny Manchanda\inst{4} \and
Girish Varma\inst{1}} 
\authorrunning{Furqan Shaik et al.}
%
\institute{IIIT Hyderabad, India \and
ServiceNow, Hyderabad
\and
Ola Krutrim, Hyderabad \and DYSL-AI, DRDO\\}

%
\maketitle              
\begin{abstract}
The ``ICPR 2024 Competition on Safe Segmentation of Drive Scenes in  Unstructured Traffic and Adverse Weather Conditions \footnote{\url{https://mobility.iiit.ac.in/safe-seg-24/index.html}\\ Prize Money for the competition was supported by iHub Data, IIIT Hyderabad.}'' served as a rigorous platform to evaluate and benchmark state-of-the-art semantic segmentation models under challenging conditions for autonomous driving. Over several months, participants were provided with the IDD-AW dataset, consisting of 5000 high-quality RGB-NIR image pairs, each annotated at the pixel level and captured under adverse weather conditions such as rain, fog, low light, and snow. A key aspect of the competition was the use and improvement of the ``Safe mean Intersection over Union (Safe mIoU)'' metric, designed to penalize unsafe incorrect predictions that could be overlooked by traditional mIoU. This innovative metric emphasized the importance of safety in developing autonomous driving systems. The competition showed significant advancements in the field, with participants demonstrating models that excelled in semantic segmentation and prioritized safety and robustness in unstructured and adverse conditions. The results of the competition set new benchmarks in the domain, highlighting the critical role of safety in deploying autonomous vehicles in real-world scenarios. The contributions from this competition are expected to drive further innovation in autonomous driving technology, addressing the critical challenges of operating in diverse and unpredictable environments.


\end{abstract}
\section{Introduction}
Semantic segmentation of driving scenes is vital for advancing autonomous vehicle technology, yet significant challenges remain, particularly in capturing rare events and adverse conditions. Optimizing segmentation models for safe driving in challenging weather and high traffic density is a critical research area at the intersection of pattern recognition, computer vision, and transportation engineering. Even though there have been many datasets dedicated to segmentation in adverse weather \cite{acdc,ithaca365,rainy_night,wilddash}, they have limited labels in structured traffic conditions only and doesn't fully capture the corner cases. State-of-the-art models achieve high accuracy under standard conditions, their performance in these corner cases remains a concern. This uncertainty poses a significant barrier to autonomous vehicles' full and safe deployment, underscoring the need for more robust and resilient models.

Safety is paramount in autonomous driving, prompting a reassessment of current evaluation metrics, such as mean Intersection over Union (mIoU). Accurately segmenting traffic participants—like vehicles and pedestrians—and roadside objects is far more critical than predicting distant objects. Additionally, the impact of mispredictions varies greatly; for example, misclassifying a car as a bus is far less critical than mistaking a car for part of the road. However, mIoU treats all mispredictions equally, failing to account for their varying implications for safety, which highlights the need for more nuanced metrics in evaluating model performance.
The robustness of a dataset is also influenced by the ambiguity of labels and the misclassification of corner cases. Addressing label ambiguity can be achieved by incorporating a hierarchical structure into the dataset labels. Hierarchical semantic segmentation offers a multi-level abstraction of visual scenes using a structured class label hierarchy, enabling a more nuanced understanding of the environment.

For this competition, we employ the IDD-AW dataset \cite{idd_aw}, which surpasses existing datasets in terms of label diversity and incorporates Near-Infrared (NIR) modality. This additional modality enhances the dataset's safety and applicability. The IDD-AW dataset features unstructured driving scenes, increasing complexity and showcasing its potential for various applications in autonomous driving and safe segmentation.

This competition makes significant contributions in several areas:
\paragraph{Main Contributions.}
    \begin{enumerate}
    \item Enhancing Autonomous Vehicle Safety: By focusing on safety in semantic segmentation, the competition aids in developing models that improve the reliability of autonomous driving, particularly under adverse weather conditions.

    \item Fostering Technological Innovation: Participants will push the boundaries of computer vision and machine learning, creating models optimized for safety in challenging environments and driving advancements in segmentation techniques.

    \item Dataset Advancement: The introduction of the IDD-AW\cite{idd_aw} dataset addresses a crucial need for diverse, challenging datasets in adverse weather, providing a valuable resource for future research in autonomous driving and computer vision.
    \end{enumerate}

\section{Competition Overview}
The central challenge addressed by the competition is optimizing the Safe mIoU metric for semantic segmentation in adverse weather conditions to ensure the safety of autonomous driving systems. Semantic segmentation is pivotal in enabling autonomous vehicles to understand their surroundings, but adverse weather conditions pose significant challenges, often leading to hazardous situations. The competition aims to develop robust models that accurately segment driving scenes even in adverse weather conditions, focusing on prioritizing safety-related incorrect predictions. This competition spans various research domains, including computer vision, machine learning, autonomous systems, and transportation engineering. Participants will leverage advanced deep-learning techniques to tackle the complexities of semantic segmentation in adverse weather scenarios.

\subsection{Dataset}
IDD-AW Dataset is an adverse weather conditions dataset captured in unstructured driving scenarios in India \footnote{\url{https://iddaw.github.io/}}. IDD-AW dataset, specifically curated to capture the challenges of unstructured traffic in adverse weather, provides a realistic test bed for evaluating model performance. Prioritize safety and robustness in model predictions. This challenge will drive innovation in autonomous driving technology by addressing critical safety concerns in real-world scenarios and prioritising safety \& robustness in model predictions. This dataset consists of 5000 RGB and NIR Image pairs. We have provided the participants with the training and validation sets for the challenge. These sets again have sub-folders with images collected in Rain, Fog, Snow and Low light. Below are the statistics for those images in Table \ref{tab2}. The competition utilizes the IDD-AW dataset, which captures highly unstructured traffic scenes in adverse weather conditions. The dataset comprises 5000 images manually selected to represent various adverse weather scenarios, including rain, fog, low light, and snow. Each RGB image also has a paired NIR image to provide image enhancement. Each image is densely annotated at the pixel level for semantic segmentation, utilizing a label set with a hierarchical structure consisting of 7 labels at level 1 and 30 at level 4 in Fig. \ref{fig:data}. IDD-AW is split into four sets corresponding to adverse weather conditions: rainy, foggy, low light, and snowy. We manually selected 1500 rainy, 1500 foggy, and 1000 low light and 1000 snowy images from the recordings for dense pixel-level semantic annotation, for a total of 5000 adverse-condition images. The dataset is further split into training and testing sets, with careful consideration to ensure that the test set comprises unseen and distinct scenes. 

\begin{table}[!ht]
    \caption{Table showing the IDD-AW datasets statistics for train, validation, and test set.}\label{tab2}
    \centering
        \begin{tabular}{lrrrrr}
        \toprule
                  & \qquad Rain & \qquad Fog & \qquad Lowlight & \qquad Snow & \qquad Total  \\
        \midrule
            Train & 1062 & 1033  & 684 & 651 & 3430 \\ 
            Val  & 120  & 154    & 100 & 101 & 475  \\
            Test & 318  & 313    & 216 & 248 & 1095 \\
            \midrule
            Total & 1500  & 1500   & 1000 & 1000 & 5000 \\ 
        \bottomrule
        \end{tabular}
        \vspace{-1em}
    \end{table}

\begin{figure}[!ht]
    \centering
    \includegraphics[width=0.75\textwidth]{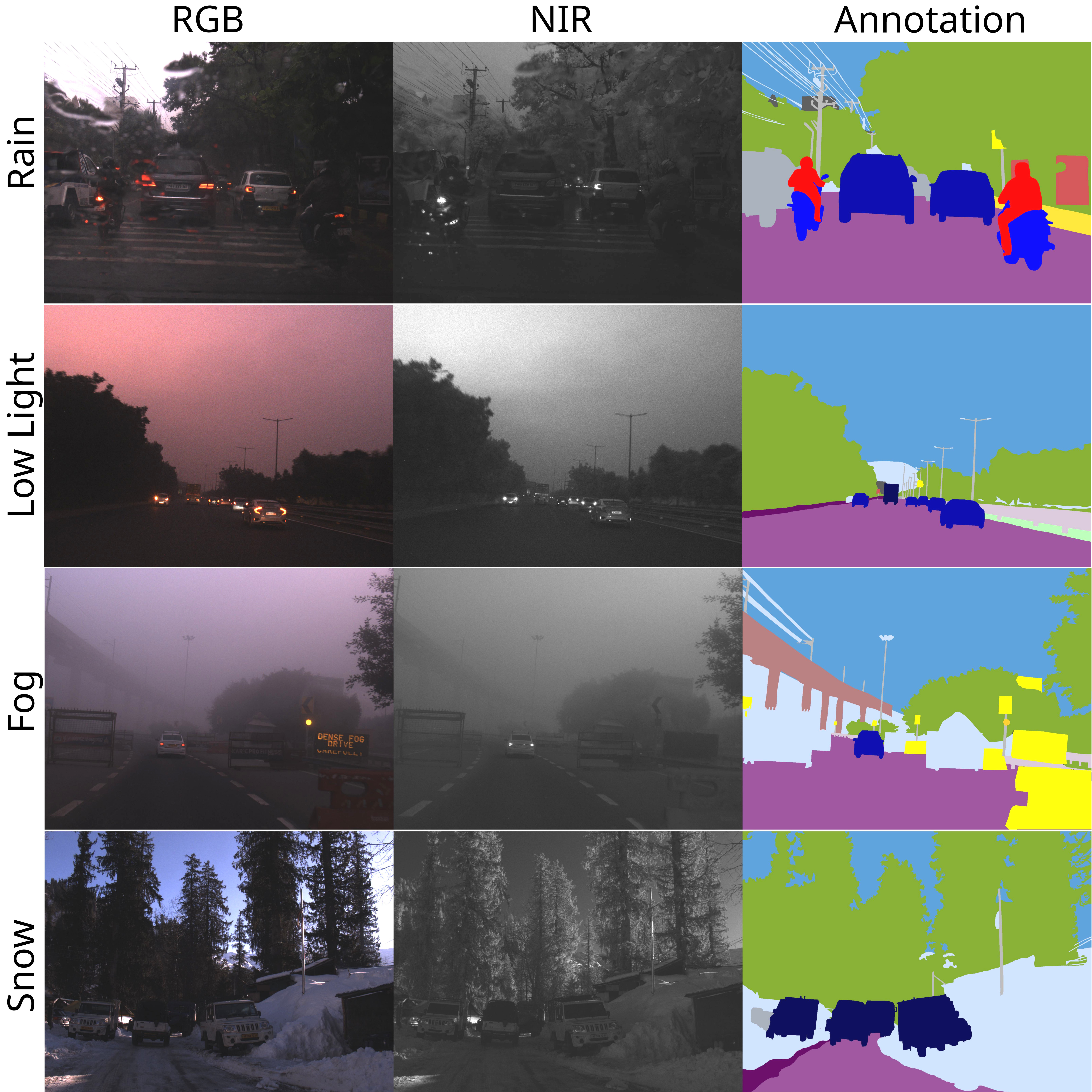}
    \caption{Examples of RGB/NIR/Annotations triplets in four adverse weather conditions from IDD-AW.} \label{fig1}
\end{figure}

\subsection{Task and Evaluation Metrics}
\subsubsection{Task:} 
mIoU (mean Intersection over Union) is a widely adopted metric for evaluating semantic segmentation tasks, as it measures segmentation quality by accounting for false positives (pixels incorrectly classified as a certain class) and false negatives (pixels of the class that are not detected). While mIoU serves well as a general-purpose metric, its limitations become evident when assessing driving scene safety. In these scenarios, accurately classifying safety-critical elements such as pedestrians, vehicles, and traffic signs is crucial, as misclassifications can have serious consequences. 

Some misclassifications are more hazardous than others—for instance, confusing one vehicle type with another is less concerning than mistaking a pedestrian or vehicle for part of the road. Unfortunately, traditional mIoU fails to reflect these safety concerns, as it treats all classes equally without considering their varying impacts on real-world driving scenarios.

We provided the participants with a task to optimize safe mIoU for the IDD-AW dataset. For this, we provided the participants with the train and validation sets of the IDD-AW dataset, each of which was divided into Rain, Fog, Low light and Snow conditions. Each condition has RGB and NIR Image pairs and semantic segmentation ground truths. Here, the participants were free to choose either just the RGB or NIR or to have some preprocessing to include both RGB and NIR Images in their training. No pre-conditions were specified or forced in this to give complete freedom in their creative solutions. 

As for the test set, we have anonymized the images to keep the evaluation fair and generic over all models. We have combined all the images into one folder, removing any relevant information about the weather conditions or the order of the images. 

The main goal of the task was to optimize the Safe mIoU (SmIoU) metric introduced in the IDD-AW Dataset Paper. The participants had to generate the submissions on the given test set,  which were then evaluated for both mIoU and SmIoU, along with SmIoU for important classes like traffic participants.


\subsubsection{Safe mIoU metric:}

\begin{figure}[!ht]
    \centering
    \includegraphics[width=0.8\textwidth]{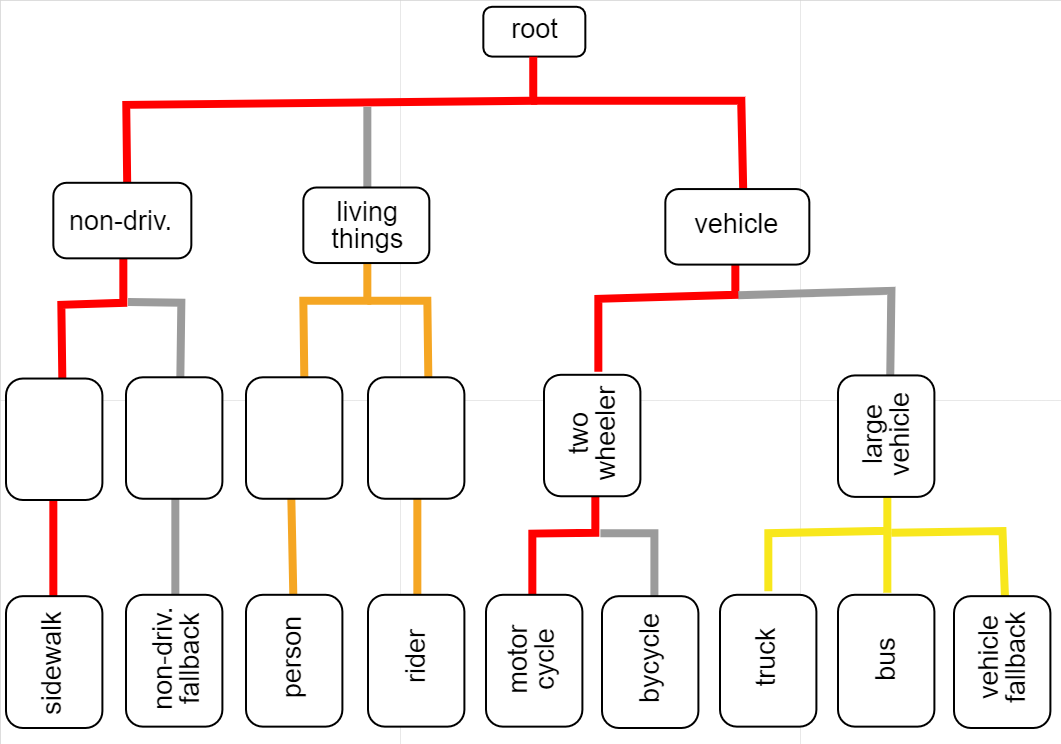}
    \caption{Example of tree distances between labels. Tree distance is used to assess the severity of mispredictions. It measures the distance between the predicted label and the correct label within the hierarchy. Red, orange, and yellow colors represent different levels of severity, with red indicating the most severe mispredictions and yellow indicating the least severe. The $\text{td} (\text{sidewalk},\text{motorcycle})=3$ since the length of the path is 6 and td is length/2. Similarly $\text{td} (\text{person},\text{rider})=2$ and $\text{td}(\text{truck},\text{bus})=1$.} 
    \label{fig:data}
\end{figure}

The essence of SmIoU lies in introducing a hierarchical penalty, a strategy that considers the semantic relationships between classes. Incorrect classifications within critical and non-critical classes classified as critical are penalized based on their distance in the class hierarchy. The \emph{tree distance} ($td$) between a pair of labels is the length of the shortest path in the class hierarchy tree divided by 2 (see Figure \ref{fig:data}). So the $td$ between person and rider is 1, while sidewalk and motorcycle are 3, denoting that the latter is a more severe incorrect prediction.

We compute safe IoUs for each class by incorporating a penalty weighted by tree distance. The final SmIoU is the mean of these safe IoUs, offering a metric for segmentation quality and safety in driving scene applications.

Let $C$ be the set of all classes at the bottom level of the hierarchy,  $d(c,s)$ be the tree distance between class $c$, $s$, and $n$ be the number of levels in the hierarchy, $\text{gt}_c$ the set of pixels in the ground truth with label $c$ and $\text{pred}_s$ the set of pixels in the prediction with label $s$. We define the following quantities:
 

\begin{equation}
    I_{c,s}^{\text{safe}} = \frac{|\text{gt}_c \cap \text{pred}_s|}{|\text{gt}_c \cup \text{pred}_c|}
    \label{eq:1}
\end{equation}

\begin{equation}
    I_{c,c} = \frac{|\text{gt}_c \cap \text{pred}_c|}{|\text{gt}_c \cup \text{pred}_c|}
    \label{eq:2}
\end{equation}

Now, we define SmIoU as follows:
\begin{equation} \label{eq:3}
 I_c^{\text{safe}} =
    \begin{cases}
    \begin{aligned}[t]
            I_{c,c} & - \sum_{s \in C , s \neq c}\frac{d(c,s)}{n}I_{c,s}^{\text{safe}} & \text{ if } c \in C_{\text{imp}} \\
            I_{c,c} & - \sum_{s \in C_{\text{imp}}}\frac{d(c,s)}{n}I_{c,s}^{\text{safe}}& \text{else.}
    \end{aligned}
    \end{cases} 
\end{equation}

\begin{equation}\label{eq:4}
    \text{SmIoU} = \frac{\sum_{c\in C} I_c^{\text{safe}}}{|C|}
\end{equation}

Note that the definition of SmIoU requires the definition of important classes $C_{\text{imp}}$. When SmIoU is mentioned without specifying $C_{\text{imp}}$, we consider it the union of traffic participants and roadside object classes.






\subsubsection{Boilerplate Code:}
As the competition was based on a new metric and a new dataset, we provided a boilerplate code \footnote{\url{https://github.com/Furqan7007/iddaw_kit/}} to give the participants a quick head start and a better understanding of the problem statement.

\subsubsection{Participation and Prizes:}

A total of 34 teams registered for the challenge, with 8 teams successfully making at least one final submission for the task. Ultimately, 26 valid final submissions were received. To encourage participation, we offered monetary prizes to the top three teams in each task, as determined by the public leaderboard on the IDD-AW ICPR Competition website. The prizes were structured as follows: \$1,000 for the best-performing team, \$500 for the second-place team, and \$300 for the third-place team. These incentives were generously provided by the iHuB-Data Foundation.

\section{Results}

\subsection{Leaderboard}

The leaderboard thoroughly describes segmentation model performance by assessing standard mean Intersec; hence, Union (mIoU) and Safe mIoU (SmIoU) metrics. With a leading SmIoU of $64.73\%$ and the corresponding mIoU of $68.32\%$, the top-performing team, KweenCoders, demonstrated the overall performance and safety of the model. They performed exceptionally well in safety-critical settings, as evidenced by their SmIoU for traffic participants (SmIoU($tp$)) was the highest at $55.07\%$. Even though SixthSenseSegmentation had the best mIoU at $68.54\%$, it couldn't perform as well at the SmIoU metric compared to the winners. This shows that both mIoU and SmIoU are not directly correlated; hence, having a greater mIoU doesn't guarantee the best safe model for segmentation. With SixthSenseSegmentation at $63.52\%$ and SemSeggers at $62.15\%$ in their respective SmIoUs, the SmIoUs on just the traffic participants were $52.08\%$ and 51.60\%. This showed a clear decline when it comes to the important classes. Though somewhat less than KweenCoders', SixthSenseSegmentation's SmIoU($tp$) of $52.08\%$ demonstrated the model's efficacy in traffic participant identification despite solely utilizing RGB data.
We also listed the IDDAW paper baselines as reported with RGB + NIR data and just the RGB-trained model, and Anidh \& Krushna bisected the two entries and came in 4th place, narrowly missing the prizes. Sanket and xmba15 teams finished further down the table. 

\begin{table}[hb]
\caption{Performance Comparison of Different Teams and Methods on RGB + NIR Image Segmentation Task, Ranked by SmIoU Scores. Here, the NIR column refers to whether the team has used NIR for Image processing. }\label{leaderboard}
    \centering
        \begin{tabular}{llrrrr}
        \toprule
        Rank   & Team Name & NIR & \quad mIoU & \quad SmIoU & \quad SmIoU($tp$) \\
        \midrule
            1 & KweenCoders & \checkmark & 68.32 & \textbf{64.73} & \textbf{55.07}\\ 
            2 & SixthSenseSegmentation & - & 68.54 & 63.52 & 52.08 \\
            3 & SemSeggers & \checkmark & 67.58 & 62.15 & 51.60\\ 
            4 & Furqan et al. RGB + NIR (baseline) & \checkmark & 67.30 & 62.26 & 51.50 \\ 
            5 & Anidh \& Krushna & - & 66.49 & 61.85 & 50.48 \\
             6 & Furqan et al. RGB (baseline) & - & 64.70 & 60.56 & 51.32 \\ 
             7 & Sanket & - &  64.82 & 60.22 & 48.86 \\
            8 & xmba15 & - & 32.18 & 17.91 & 2.10 \\
        \bottomrule
        \end{tabular}
    
\end{table}

\subsection{Classwise Comparison}

\begin{table*}[t]
\caption{Comparison of class-wise labels for important classes between mIoU vs SmIoU for top 3 participants as well as the current SOTA}\label{miou_vs_safemiou_classwise}
\centering
\resizebox{\textwidth}{!}{
\begin{tabular}{@{}llccrrrrcrrrrcrrcc@{}} \toprule
{Team}
&
    Metric 
    & \rotatebox{90}{road}
    &\rotatebox{90}{\begin{tabular}[c]{@{}l@{}}drivable\\ fallback\end{tabular} }
    &\rotatebox{90}{sidewalk}
    &\rotatebox{90}{person}
    &\rotatebox{90}{rider}
    &\rotatebox{90}{bike} 
    &\rotatebox{90}{bicycle}
    &\rotatebox{90}{rickshaw}
    &\rotatebox{90}{car}
    &\rotatebox{90}{truck}
    &\rotatebox{90}{bus}
    &\rotatebox{90}{\begin{tabular}[c]{@{}l@{}}vehicle\\ fallback\end{tabular}}
    &\rotatebox{90}{curb}
    &\rotatebox{90}{wall}
    &\rotatebox{90}{\begin{tabular}[c]{@{}l@{}}traffic\\ sign\end{tabular}}
    &\rotatebox{90}{\begin{tabular}[c]{@{}l@{}}traffic\\ light\end{tabular}} \\
    \midrule
    {KweenCoders}&
    mIoU & 95 & 48 & 55 & 83 & 80 & 75 & 17 & 85 & 85 & 74 & 73 & 50 & 80 & 56 & 62 & 62  \\
    
    &SmIoU & 94 & 48 & 54 & 75 & 70 & 63 & -6 & 82 & 82 & 62 & 66 & 33 & 79 & 56 & 62 & 61\\
    \midrule
    
    {SixthSenseSegmentation}&
    mIoU & 95 & 53 & 50 & 78 & 77 & 73 & 12 & 84 & 86 & 77 & 72 & 49 & 81 & 54 & 62 & 57 \\
    
    &SmIoU & 95 & 52 & 50 & 67 & 66 & 60 & -20 & 79 & 83 & 72 & 63 & 32 & 80 & 53 & 62 & 56 \\
    \midrule
    
    {SemSeggers}&
    mIoU & 94 & 43 & 46 & 81 & 78 & 74 & 16 & 79 & 83 & 61 & 77 & 43 & 78 & 50 & 60 & 60 \\
    
    &SmIoU & 94 & 42 & 46 & 72 & 67 & 62 & -10 & 71 & 79 & 44 & 73 & 26 & 77 & 50 & 59 & 59 \\
    \midrule
    {IDD-AW Paper (Baseline)}&
    mIoU & 95 & 51 & 48 & 76 & 72 & 68 & 5 & 83 & 85 & 74 & 76 & 45 & 78 & 52 & 58 & 52 \\
    
    &SmIoU & 92 & 32 & 16 & 64 & 58 & 52 & -22 & 77 & 81 & 68 & 70 & 21 & 69 & 32 & 40 & 27 \\
    \bottomrule
  \end{tabular}
}
\vspace{-1em}

\end{table*}

Classwise Comparison details how each participant's model has performed in mIoU and SmIoU across various classes. We can see that the performance of all the class participants, such as road, car, curb, and traffic signs, are almost similar. Compared with the IDD-AW Paper baseline, the top 3 participants performed very well on SmIoU in many classes, such as on the sidewalk, bike, person, rider, traffic sign, and traffic light. Also, SixthSenseSegmentation, the 2nd place winner, performed better than KweenCoders in multiple classes, such as drivable fallback, car, truck, and curb. However, the winners, KweenCoders, have performed better in critical classes like a sidewalk, person rider, bicycle, rickshaw, vehicle fallback, and traffic light. This shows why the KweenCoders team reported better SmIoU across traffic participants than other teams, as shown in table \ref{tab2}. 

\subsection{Qualitative Comparison}
\begin{figure}[ht]
    \centering
    \includegraphics[width=0.95\linewidth]{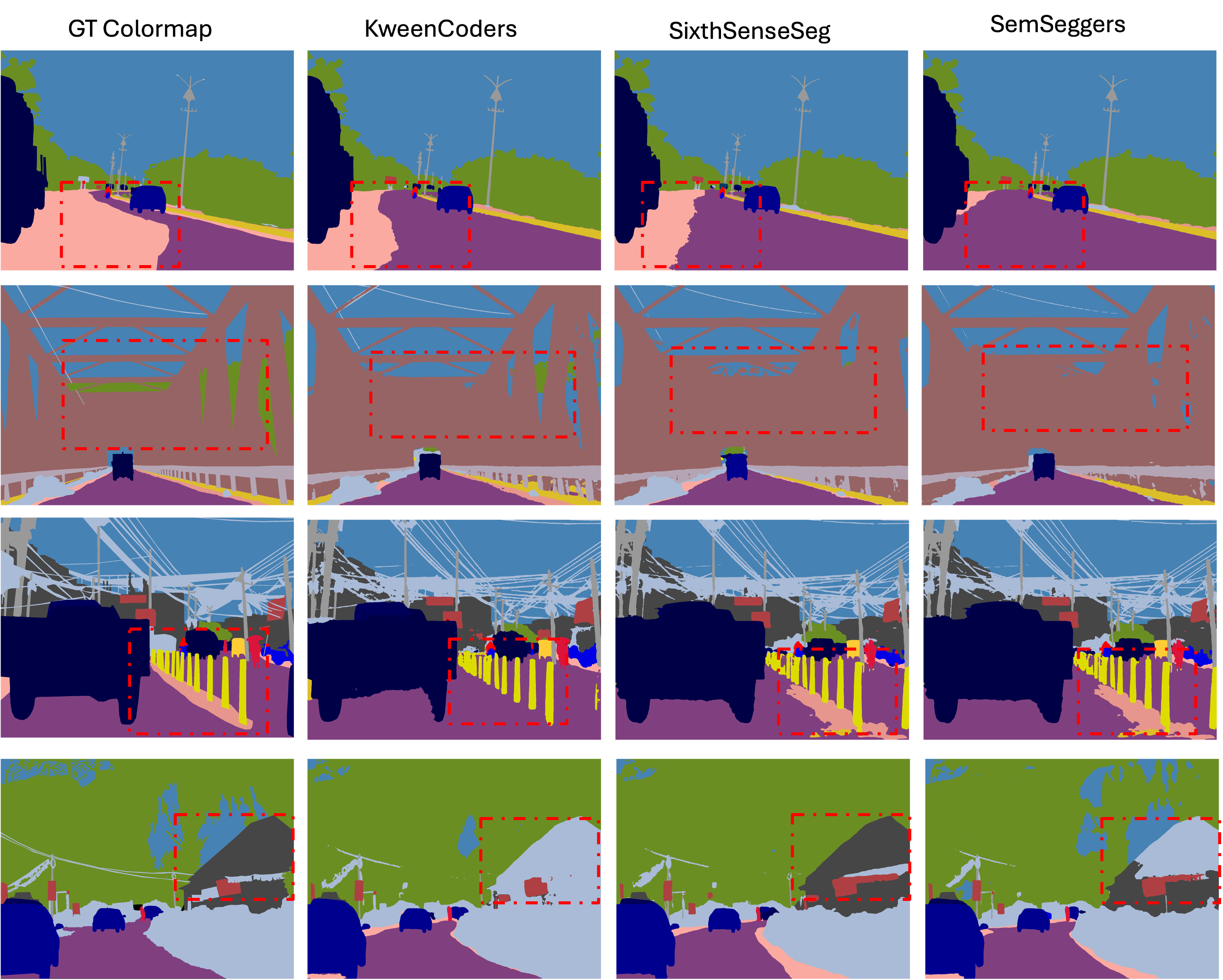}
    \caption{Qualitative Comparision of predicted colourmaps between ground truth and top 3 participants. The pixels inside the red highlighted boxes show the major differences between the ground truth and all 3 approaches.}
    \label{fig:colormap_comp}
\end{figure}

In Fig \ref{fig:colormap_comp}, we have shown several examples from each participant when compared with the ground truth. In the first image, we can see that the road is not well segmented in either of the predictions, with KweenCoders being somewhat close, but SemSeggers also predicts the whole sidewalk as a road label. In the 2nd row, the bridge is not well segmented in the last column, whereas it is better in the first 2 participants. This is because the first 2 teams pre-trained their model on the IDD Dataset, which has the bridge label, whereas the CityScapes pre-trained models don't have a bridge class. In the 3rd row, the sidewalk is not segmented in the KweenCoders prediction. but the poles are well-segmented. In SixthSenseSegmentation and SemSeggers, they have predicted the sidewalk but find it difficult to predict the poles correctly. In the last Image, the fallback background is mispredicted by the KweenCoders, but it's well predicted by the 2nd-place participants. This shows that the participants have performed well across various classes and have gotten good predictions, but even though the SmIoU scores and the mIoU scores might be similar or very close, we can see clear distinctions in their predictions. 

\vspace{-1em}

\subsection{Histogram of Errors Comparison}
\begin{figure}[ht]
    \centering
    \includegraphics[width=0.9\textwidth]{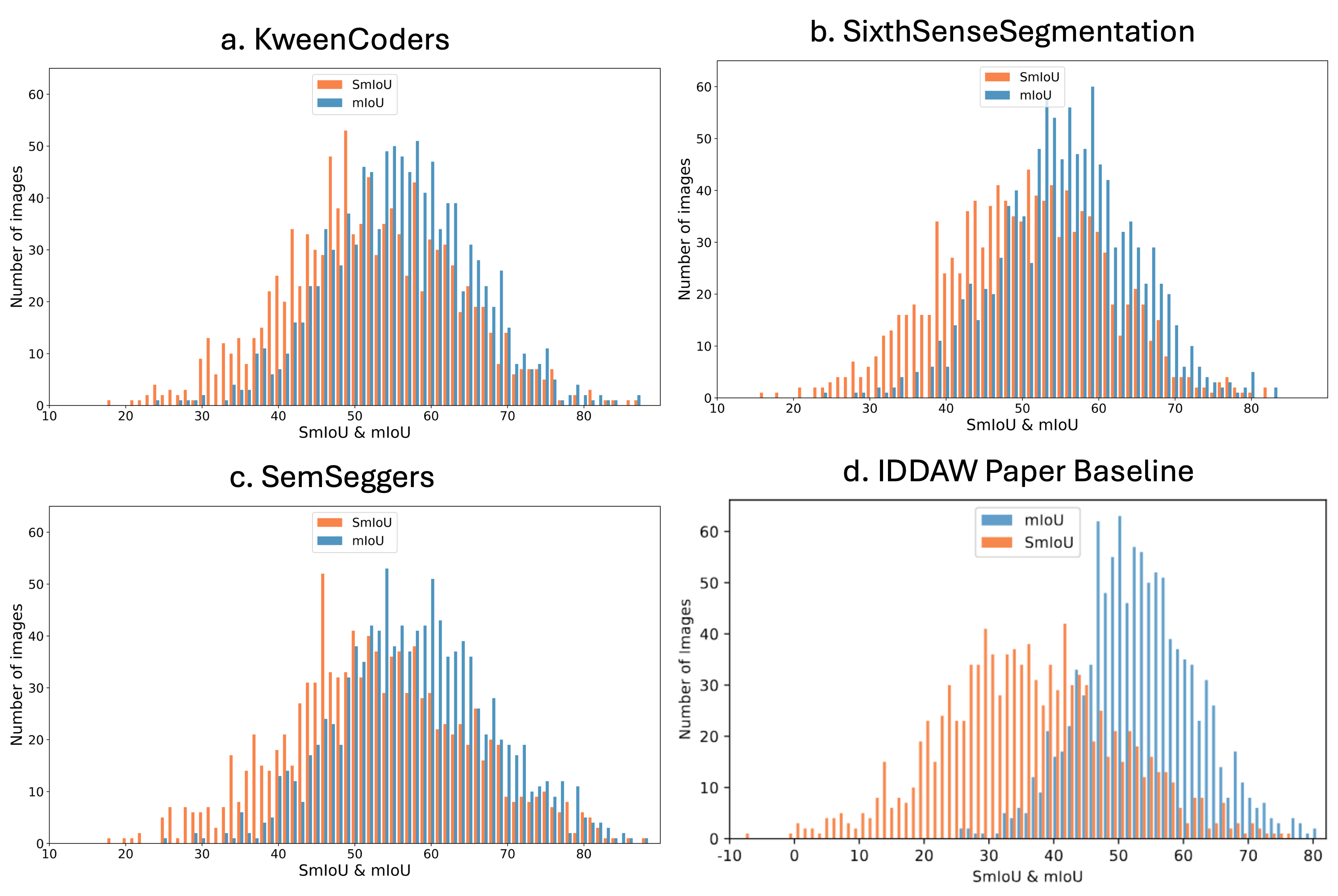}
    \caption{Distribution of Safe mIoU (\%) vs mIoU (\%) for top 3 participants and the IDD-AW paper baseline on IDD-AW test set. y-axis represents the number of images with a particular value of SmIoU/mIoU.}
    \label{fig:smiou_total}
\end{figure}

We have plotted the histograms for the top 3 participants for the SmIoU vs. mIoU comparison and the IDD-AW paper baseline plot. This was done to know the values of how mIoU and SmIoU were distributed across several images. Here, we can see that all the top 3 participants have much more overlapping mIoU vs SmIoU values when compared with the original IDD-AW paper. This shows that the gap between mIoU and SmIoU was brought closer by this competition, which was one of the main objectives of this challenge. In the first plot, many images have SmIoU and mIoU between $45-70\%$. This is one of the main reasons they have achieved such high performance over the whole dataset. Even though their mIoU is not the highest, bridging this gap has helped them boost their SmIoU score better than the other participants. In the 2nd plot, the SixthSenseSegmentation has improved mIoU as seen in the plot. However, they have fewer images in the same buckets for SmIoU. The 3rd place participants, SemSeggers, also have good overlap, but most of their images have SmIoU  $< 50\%$ 



\section{Conclusion}
The insights obtained from the Safe Segmentation competition can be summarized as follows:

\begin{enumerate}
    \item The methods introduced in this competition have significantly advanced the state of the art in road image segmentation, paving the way for practical applications in autonomous driving.
    \item The prominence of RGB + NIR fused images, which secured 2 of the top 3 positions, underscores the value of Near-Infrared (NIR) images in enhancing safety, particularly under adverse weather conditions. 
    \item The continued strong performance of Transformer architectures and Transfer Learning highlights their effectiveness, while SmIoU loss-based approaches and innovative model training techniques reveal promising avenues for architectural design, demonstrating robust reconstruction capabilities.
    \item Scalability remains a critical factor for image segmentation and autonomous driving communities, emphasizing the growing need for expansive datasets and the integration of RGB-NIR image fusion methods.
    \item The autonomous driving and image segmentation communities must address several critical challenges, including multi-modality, interpretability, causal confusion, robustness, and the development of comprehensive world models, among others. 
\end{enumerate}

\section{Acknowledgements} First and foremost, we would like to thank our sponsor, i-HuB Data Foundation, which made the widespread reach of the competition possible and the allocation of prize money, contributing to the incentive for our competition. Also, we would like to thank all the participants in this competition who have been the real protagonists of this event.

\section{Competition Teams and Methods}
In Table \ref{leaderboard}, the results for the task are mentioned. As we can see from Table \ref{tab2}, the participants managed to improve over the initial proposed baseline model. They also improved the current SOTA benchmark on the IDD-AW dataset for both mIoU and Safe mIoU metrics. We describe the winning solutions and provide the proposed image processing methods.

\subsection{Participants}

\subsubsection{1st Classified - KweenCoders (Aiswarya Maturi):}
The participants' approach consists of three key steps: 
\begin{enumerate}
    \item Data pre-processing: This step involves fusing RGB and NIR images to enhance the RGB images, improving visibility and feature extraction (see Fig \ref{fig:fusion_net}).
    \item Loss Function Update: The loss function has been refined to achieve better performance on the safe mIoU metric.
    \item Training: The SegFormer model \cite{segformer} is employed, initially pre-trained on the IDD-AW dataset and subsequently fine-tuned on the IDD-AW Dataset.
\end{enumerate}

\subsubsection{Data pre-processing:}
\begin{figure}[ht]
    \centering
    \includegraphics[width=0.9\linewidth]{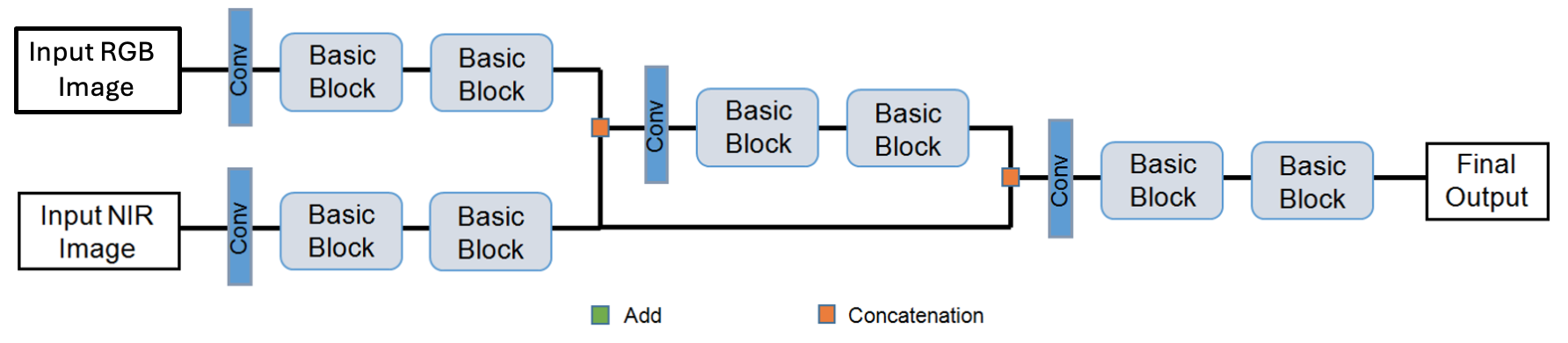}
    \caption{Fusion net architecture.} \label{fig:fusion_net}
    \vspace{-2em}
\end{figure}

We fuse RGB and NIR images using the FusionNet\cite{fusionnet} method for the data preprocessing. In this, RGB and NIR Images are passed through separate CNN networks, and then those features are passed through a final CNN network, which gives the final enhanced RGB Image. The FusionNet architecture is shown in the Fig. \ref{fig:fusion_net}. The code for both the training and preprocessing is provided here\footnote{\url{https://github.com/maturiaiswarya/SegFormer/}}. 

\subsubsection{Loss Function:}
We used a combination of the Cross-Entropy Loss and Dice Loss for the loss function. 

\begin{equation}
    \text{CrossEntropyLoss} = -\frac{1}{N} \sum_{i=1}^{N} \sum_{c=1}^{C} y_{i,c} \log(p_{i,c})
\end{equation}

\begin{equation}
    \text{DiceLoss} = 1 - \frac{2 \sum_{i=1}^{N} p_i y_i}{\sum_{i=1}^{N} p_i + \sum_{i=1}^{N} y_i}
\end{equation}

\begin{equation}
    \text{Loss} = 0.5 \times \text{CrossEntropyLoss} + 0.5 \times \text{DiceLoss}
\end{equation}

\subsubsection{Training Approach:}
Following the preprocessing phase, the enhanced RGB images are input into the SegFormer model \cite{segformer}. SegFormer is highly effective in semantic segmentation tasks due to its efficient hierarchical transformer architecture, which captures multi-scale features without relying on positional encoding. The model employs a combination of cross-entropy and dice loss in the MLP decoder to optimize segmentation performance. SegFormer’s architecture includes a series of transformer blocks that progressively increase feature map resolution, each comprising an efficient self-attention layer and a feed-forward network. The decoder integrates multi-level features from the encoder through simple linear layers. SegFormer's design enables it to perform robustly even with limited data, and it can be trained from scratch while maintaining scalability. Its balance between performance and efficiency, as demonstrated in various benchmarks, makes it a leading choice for semantic segmentation.

Training Procedure: The participants initially trained the model on the IDD dataset for 150K iterations, achieving a good mean Intersection over Union (mIoU). They then fine-tuned this model on the IDD-AW dataset for 160K iterations. The participants conducted multiple runs to prevent overfitting, adjusting batch sizes and intermediate checkpoints. All training was performed using two NVIDIA V100 GPUs.


\subsubsection{2nd Classified - SixthSenseSegmentation (Harshit Kumar Sankhla, Dibyendu Ghosh):}

The participants focused on enhancing semantic segmentation performance under challenging conditions. Their solution is built on the Mask2Former\cite{mask2former}  model, a universal segmentation architecture developed by Facebook AI Research.

\subsubsection{Architecture:}
\begin{figure}
    \vspace{-1em}
    \centering
    \includegraphics[width=0.5\linewidth]{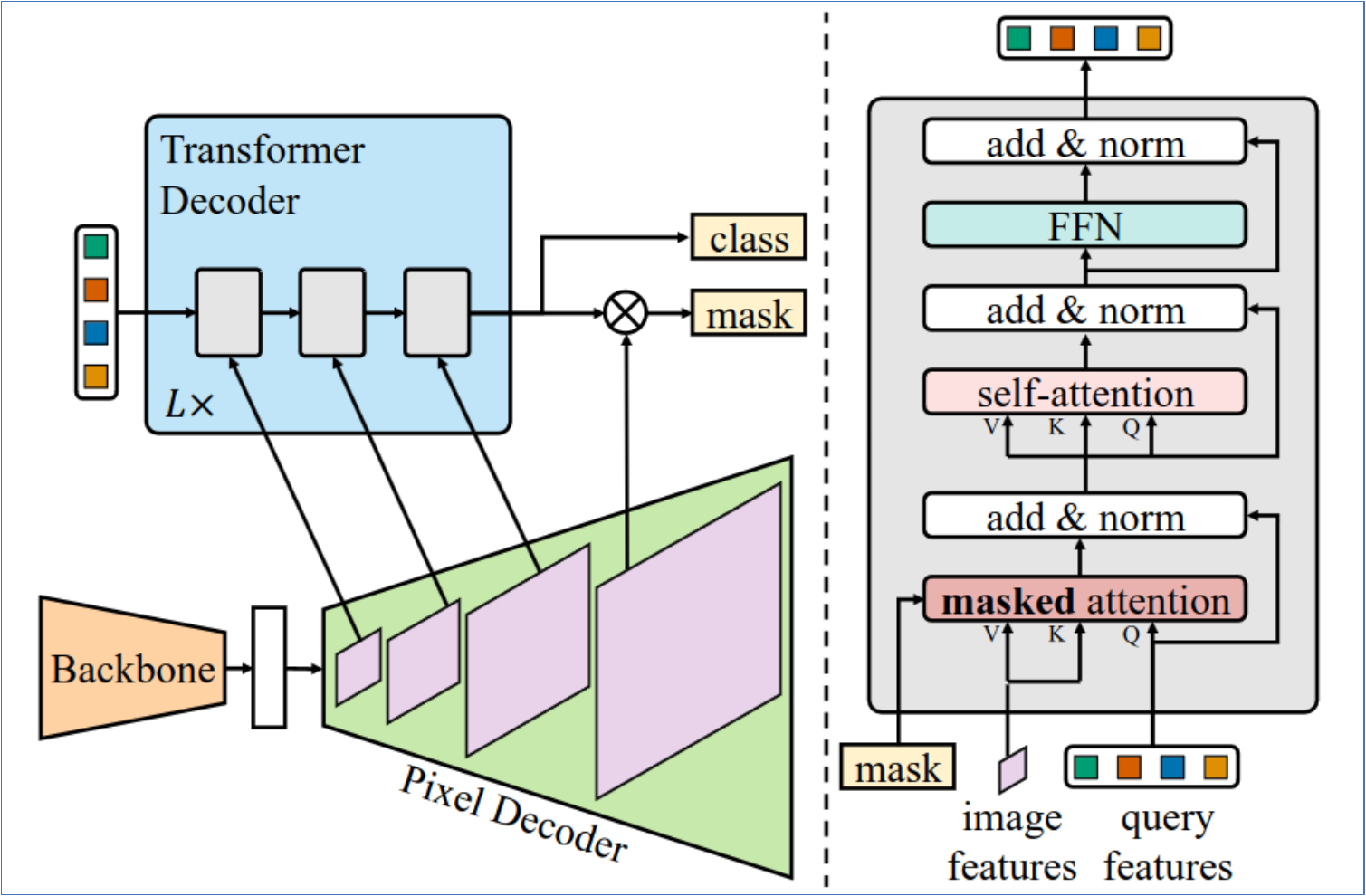}
    \caption{Mask2Former architecture: adopts the MaskFormer\cite{maskformer} backbone, a pixel decoder and a Transformer decoder with masked attention.}
    \label{fig:mask2former}
    \vspace{-1em}
\end{figure}

Mask2Former utilizes the same meta-architecture as MaskFormer \cite{maskformer} but with a novel Transformer decoder that enhances the model's ability to segment images (see Fig. \ref{fig:mask2former}). The key innovation in Mask2Former is the masked attention operator, which restricts cross-attention to the foreground region of the predicted mask for each query, rather than the entire feature map. Our implementation uses a Swin-L backbone pre-trained on ImageNet, comprising 216 million parameters. 

\vspace{-1em}
\subsubsection{Training strategy:}
The participants follow a 3-step training strategy to maximize accuracy for the task:
\begin{enumerate}
    \item They have selected a pre-trained network on CityScapes\cite{cityscapes}, readily available in the model zoo. This serves as a good starting point for fine-tuning further.
    \item Next, they fine-tune the model on the IDD (Indian Driving Dataset) \cite{idd} to adjust the model to Indian driving scenarios. This improves the model’s reasoning towards objects and structures unique to the Indian driving environment. For approximately two days, they trained their model for 60k iterations on NVIDIA DGX A100 GPUs.
    
    \item They then fine-tune the last 4 layers of the model on the IDD-AW dataset as the final step. They freeze the initial layers to ascertain consistency in the syntactic features learnt in the early layers from the previous datasets and robustify it from noise due to adverse weather conditions. This stage of training, which consists of 90k steps, takes approximately two days on NVIDIA DGX A100 GPUs.
    
\end{enumerate}

Additionally, they have incorporated the following steps to improve the overall training pipeline and accuracy 
\begin{enumerate}
    \item Normalized the images by pixel mean and standard deviation for each weather sub-category.
    \item Incorporated linear learning rate warm-up to reduce the primacy effect of the dataset and regularize the model.
    \item Set batch size relatively higher for lDD (10k images) and lower for IDD-AW (3k images).
\end{enumerate}

Evaluation: The model demonstrated strong performance across various classes in the IDD-AW dataset, achieving a mean Intersection over Union (mIoU) of 69.51 on the validation dataset provided to the participants. The model's ability to adapt to adverse weather conditions was validated through multiple runs and careful checkpointing to prevent overfitting.

\subsubsection{3rd Classified - SemSeggers (Anshuman Majumdar, Srikanth Vidapanakal):}

\begin{figure}
    \centering
    \includegraphics[width=0.9\linewidth]{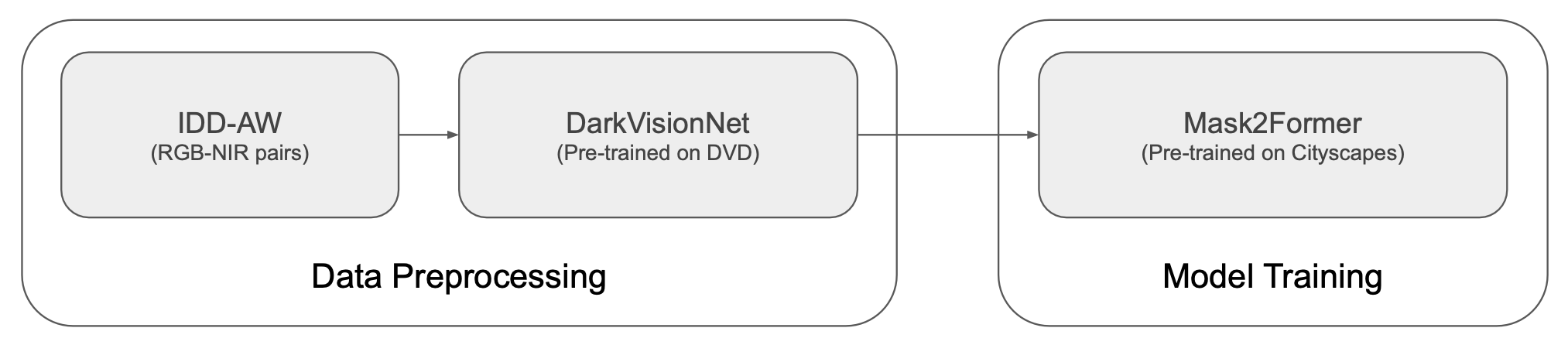}
    \caption{Flow diagram for our (SemSeggers) proposed approach.}
    \label{fig:safe_seg_01}
    \vspace{-3em}
\end{figure}

The method is comprised of two steps parts \ref{fig:safe_seg_01} - Image Smoothing using RGB-NIR pairs through a pre-trained DarkVisionNet \cite{dvn} inference pipeline on IDD-AW, followed by fine-tuning a pre-trained Mask2Former\cite{mask2former} Semantic Segmentation network.

\subsubsection{Image Smoothing:}
DarkVisionNet (DVN)\cite{dvn} obtains high-quality low-light images without the visual artefacts from RGB-NIR pairs. It enhances the low-light noisy colour (RGB) image through rich, detailed information in the corresponding near-infrared (NIR) image. The entire dataset was smoothed using a pre-trained DVN model on the DVD dataset

After image smoothing, certain discolouration and artefacts were introduced by a pre-trained DVN model. However, this does not affect the semantic segmentation model training based on experiments. The participants also observed that DVN significantly reduces pixel-level noise.

\subsubsection{Training details:}
They have used the pre-trained model Mask2Former on the Cityscapes\cite{cityscapes} dataset and fine-tuned it on the IDD-AW dataset. Initially, the experiments were configured to run for 90k iterations. Safe mIoU was not used as a loss function during training or experimentation. It is observed that the training reached a plateau, and an early stoppage was performed at $65k$ iterations to prevent overfitting.
Training Mask2Former on IDD-AW train data, early stoppage at 65k steps, and training time of 1 day, 3 hours on 2 x NVIDIA A6000 GPUs.

The model's training was carefully monitored, with early stopping employed after 65k iterations to prevent overfitting as the training loss plateaued. The resulting model demonstrated significant robustness in handling the challenging conditions presented by the IDD-AW dataset.

\subsubsection{Observations:} Smoothed images show that certain discolouration and artefacts are introduced by a pre-trained DVN model. However, this doesn’t affect the semantic segmentation model training based on experiments. The trained model seems to properly capture the semantics of important objects.

%
%
%
\bibliographystyle{splncs04}
%

\end{document}